\documentclass[letterpaper, 10 pt, conference]{ieeeconf}  
\IEEEoverridecommandlockouts
\let\labelindent\relax
\usepackage{cite}
\usepackage{amsmath}
\usepackage{amssymb}
\DeclareMathOperator*{\argmax}{arg\,max}

\usepackage{wrapfig}
\usepackage{booktabs}
\usepackage{multicol}
\usepackage{multirow}
\usepackage{tabularx}
\usepackage{diagbox}
\usepackage{graphicx}
\usepackage{subcaption}
\usepackage{array}
\usepackage{cleveref}
\usepackage{xcolor}
\usepackage{tikz}
\usepackage{verbatim}
\usepackage{listings}
\definecolor{mygreen}{RGB}{28,172,0} 
\definecolor{mylilas}{RGB}{170,55,241}

\usepackage{matlab-prettifier}
\usepackage{color}
\usepackage[shortlabels]{enumitem}

\usepackage{algorithm}
\usepackage[noend]{algpseudocode}
\makeatletter
\def\BState{\State\hskip-\ALG@thistlm}
\makeatother
\newcommand{\E}{\mathbb{E}}

\newcommand{\tp}{\mathsf{T}}

\newcommand{\R}{\mathbb{R}}

\newcommand{\st}{\text{subject to}}

\author{Tao Li$,^{1}$$^*$ Haozhe Lei$,^{1}$$^*$ and Quanyan Zhu$^{1}$ 
\thanks{$^*$The authors contributed equally to this work.}\thanks{$^{1}$The authors are with the Department of Electrical and Computer Engineering, New York University, Brooklyn, NY, 11201, USA.
        {\tt\small \{tl2636, hl4155, qz494\}@nyu.edu}. This work is partially supported by grant ECCS-1847056 from National Science Foundation (NSF).}%
}
\begin{document}

\title{\LARGE\bf Self-Adaptive Driving in Nonstationary Environments through  Conjectural Online Lookahead Adaptation
}

\maketitle

\begin{abstract}
Powered by deep representation learning, reinforcement learning (RL) provides an end-to-end learning framework capable of solving self-driving (SD) tasks without manual designs. However, time-varying nonstationary environments cause proficient but specialized RL policies to fail at execution time. For example, an RL-based SD policy trained under sunny days does not generalize well to rainy weather. Even though meta learning enables the RL agent to adapt to new tasks/environments, its offline operation fails to equip the agent with online adaptation ability when facing nonstationary environments. This work proposes an online meta reinforcement learning algorithm based on the \emph{conjectural online lookahead adaptation} (COLA). COLA determines the online adaptation at every step by maximizing the agent's conjecture of the future performance in a lookahead horizon.  Experimental results demonstrate that under dynamically changing weather and lighting conditions, the COLA-based self-adaptive driving outperforms the baseline policies regarding online adaptability. A demo video, source code, and appendixes are available at {\tt https://github.com/Panshark/COLA}  
\end{abstract}

\section{Introduction}
Recent breakthroughs from machine learning \cite{ross11dagger,mnih2015DQN,tao_confluence} have spurred wide interest and explorations in learning-based self driving (SD) \cite{kiran21drl_ad}. Among all the endeavors, end-to-end reinforcement learning \cite{chen21end2end} has attracted particular attention. Unlike modularized approaches, where different modules handle perception, localization, decision-making, and motion control, end-to-end learning approaches aim to output a synthesized driving policy from raw sensor data. 

However, the limited generalization ability prevents RL from wide application in real SD systems.  To obtain a satisfying driving policy, RL methods such as Q-learning and its variants \cite{mnih2015DQN,tao_multiRL}  or policy-based ones \cite{sutton_PG,Tao_blackwell} require an offline training process.  Training is performed in advance in a stationary environment, producing a policy that can be used to make decisions at execution time in the same environments as seen during training. However, the assumption that the agent interacts with the same stationary environment as training time is often violated in practical problems. Unexpected perturbations from the nonstationary environments pose a great challenge to existing RL approaches, as the trained policy does not generalize to new environments \cite{padakandle21nonstationary}. 

\begin{figure}
    \centering
   \includegraphics[width=0.4\textwidth]{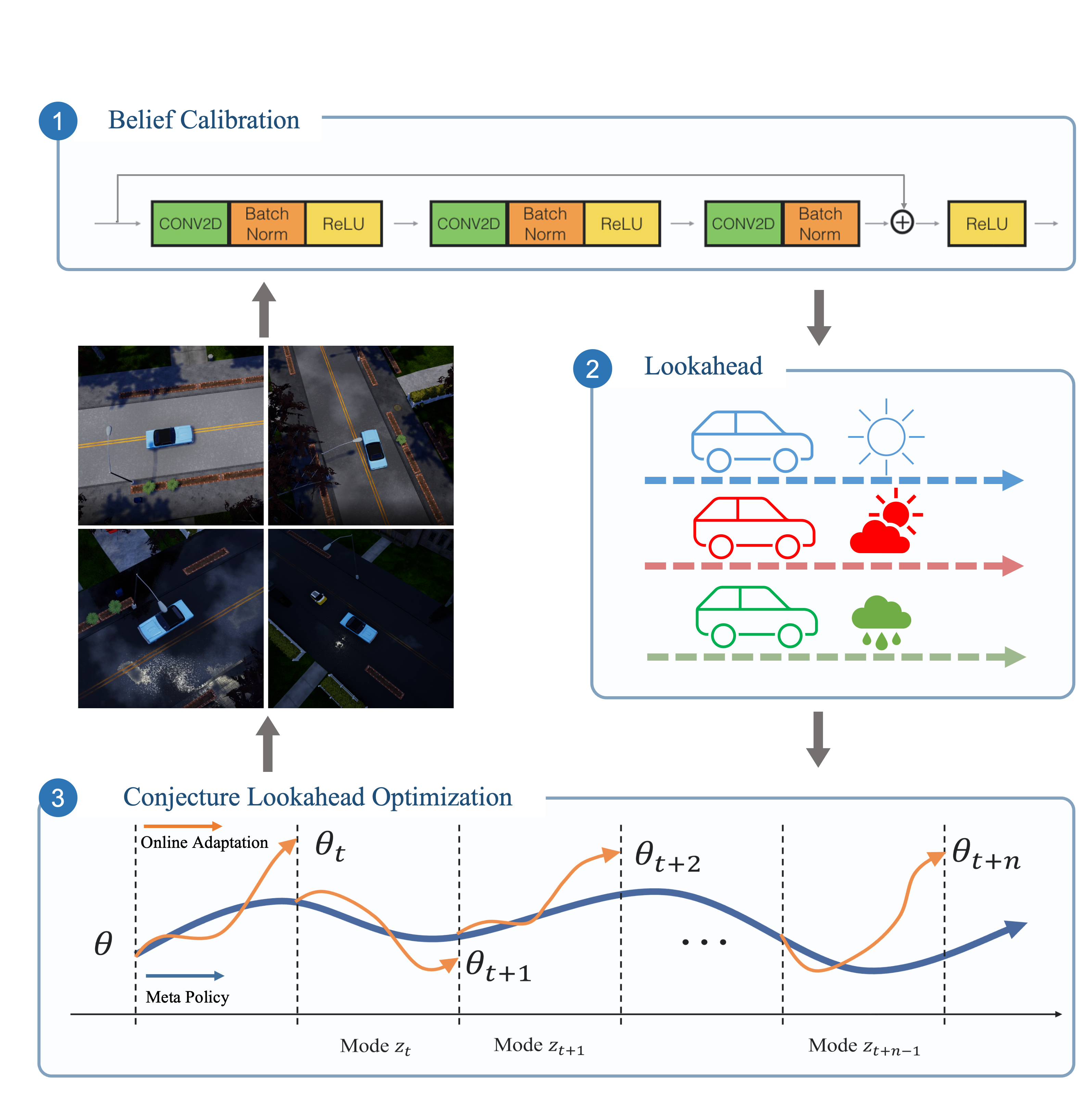}
    \caption{An illustration of conjectural online lookahead adaptation. When driving in a changing environment, the agent first uses a residual neural network and bayesian filtering to calibrate its belief at every time step about the hidden mode. Based on its belief, the agent conjectures its performance in the future within a lookahead horizon. The policy is adapted through conjectural lookahead optimization, leading to a suboptimal (empirically) online control.}
    \label{fig:cola}\vspace{-6mm}
\end{figure}

To elaborate on the limited generalization issue, we consider the vision-based lane-keeping task under changing weather conditions shown in \Cref{fig:cola}. The agent needs to provide automatic driving controls (e.g., throttling, braking, and steering) to keep a vehicle in its travel lane, using only images from the front-facing camera as input. The driving testbed is built on the CARLA platform \cite{carla}. As shown in \Cref{fig:cola}, different weather conditions create significant visual differences, and a vision-based policy learned under a single weather condition may not generalize to other conditions. As later observed in one experiment [see \Cref{fig:poor}], a vision-based SD policy trained under the cloudy daytime condition does not generalize to the rainy condition. The trained policy relies on the solid yellow lines in the camera images to guide the vehicle. However, such a policy fails on a rainy day when the lines are barely visible.        

The challenge of limited generalization capabilities has motivated various research efforts. In particular, as a learning-to-learn approach, meta learning \cite{meta_survey} stands out as one of the well-accepted frameworks for designing fast adaptation strategies. Note that the current meta learning approaches primarily operate in an offline manner. For example, in model-agnostic meta learning (MAML) \cite{finn2017model}, the meta policy needs to be first trained in advance. When adapting the meta policy to the new environment in the execution time, the agent needs a batch of sample trajectories to compute the policy gradients to update the policy. However, when interacting with a nonstationary environment online, the agent has no access to prior knowledge or batches of sample data. Hence, it is unable to learn a meta policy beforehand. In addition, past experiences only reveal information about previous environments. The gradient updates based on past observations may not suffice to prepare the agent for the future. In summary, the challenge mentioned above persists when the RL agent interacts with a nonstationary environment in an online manner. 

\textbf{Our Contributions} To address the challenge of limited online adaptation ability, this work proposes an online meta reinforcement learning algorithm based on \emph{conjectural online lookahead adaptation} (COLA). Unlike previous meta RL formulations focusing on policy learning, COLA is concerned with learning meta adaptation strategies online in a nonstationary environment. We refer to this novel learning paradigm as online meta adaptation learning (OMAL). Specifically, COLA determines the adaptation mapping at every step by maximizing the agent's conjecture of the future performance in a lookahead horizon. This lookahead optimization is approximately solved using off-policy data, achieving real-time adaptation. A schematic illustration of the proposed COLA algorithm is provided in \Cref{fig:cola}.

In summary, the main contributions of this work include that 1) we formulate the problem of learning adaptation strategies online in a nonstationary environment as online meta-adaptation learning \eqref{eq:omal}; 2) we develop a real-time OMAL algorithm based on conjectural online lookahead adaptation (COLA); 3) experiments show that COLA equips the self-driving agent with online adaptability, leading to self-adaptive driving under dynamic weather.
\begin{figure}[H]
    \centering
    \includegraphics[width=0.4\textwidth]{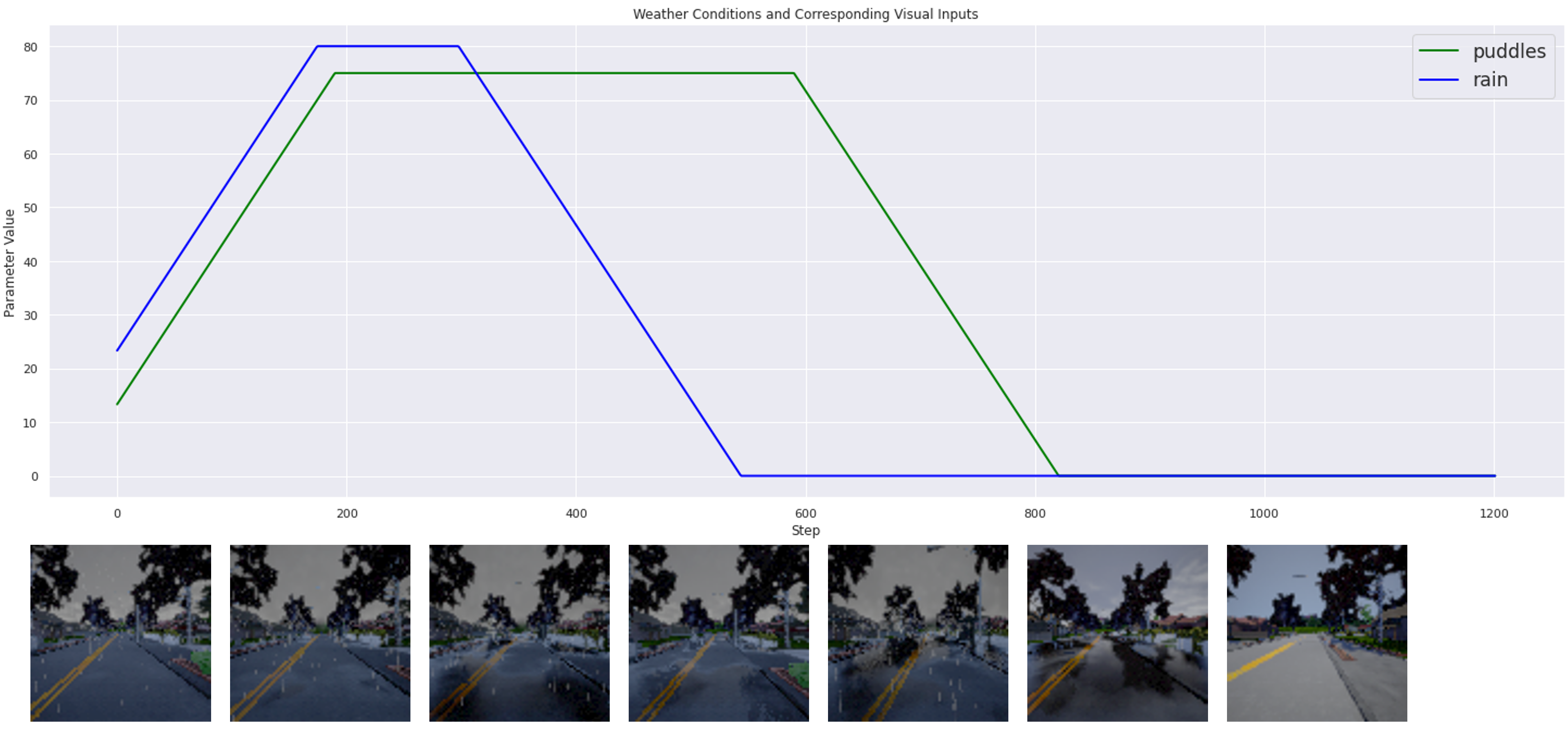}
    \caption{An example of lane-keeping task in an urban driving environment under time-varying weather conditions. Dynamic weather is realized by varying three weather parameters: cloudiness, rain, and puddles.  Different weather conditions cause significant visual differences in the low-resolution image. }
    \label{fig:image_input}\vspace{-2mm}
\end{figure}
\section{Self-Driving in Nonstationary Environments: Modeling and Challenges}\label{sec:model}
We model the lane-keeping task under time-varying weather conditions shown in \Cref{fig:image_input} as a Hidden-Mode Markov Decision Process (HM-MDP) \cite{choi2000hidden}. The state input at time $t$ is a low-resolution image, denoted by $s_t\in \mathcal{S}$. Based on the state input, the agent employs a control action $a_t\in\mathcal{A}$, including acceleration, braking, and steering. The discrete control commands used in our experiment can be found in Appendix \Cref{app:setup}. The driving performance at $s_t$ when taking $a_t$ is measured by a reward function $r_t=r(s_t,a_t)$.

Upon receiving the control commands, the vehicle changes its pose/motion and moves to the next position. The new surrounding traffic conditions captured by the camera serve as the new state input subject to a transition kernel $P(s_{t+1}|s_t,a_t;z_t)$, where $z_t\in \mathcal{Z}$ is the environment mode or latent variable hidden from the agent, corresponding to the weather condition at time $t$. The transition $P(s_{t+1}|s_t,a_t;z_t)$ tells how likely the agent is to observe a certain image $s_{t+1}$ under the current weather conditions $z_t$. CARLA simulator controls the weather condition through three weather parameters: cloudiness, rain, and puddles, and $z_t$ is a three-dimensional vector with entries being the values of these parameters.

In HM-MDP, the hidden mode shifts stochastically according to a Markov chain $p_z(z_{t+1}|z_t)$ with initial distribution $\rho_z(z_1)$. As detailed in the Appendix \Cref{app:setup}, the hidden mode (weather condition) shifts in our experiment are realized by varying three weather parameters subject to a periodic function. One realization example is provided in \Cref{fig:image_input}. Note that the hidden mode $z_t$, as its name suggested, is not observable in the decision-making process. Let $\mathcal{I}_t=\{s_t,a_{t-1},r_{t-1}\}$ be the set of agent's observations at time $t$, referred to as the information structure \cite{tao_info}. Then, the agent's policy $\pi$ is a mapping from the past observations $\cup_{k=1}^t \mathcal{I}_k$ to a distribution over the action set $\Delta(\mathcal{A})$. 

\paragraph{Reinforcement Learning and Meta Learning} Standard RL concerns stationary MDP, where the mode remains unchanged (i.e., $z_t=z$) for the whole decision horizon $H$. Due to stationarity, one can search for the optimal policy within the class of Markov policies \cite{puterman_mdp}, where the policy $\pi:\mathcal{S}\rightarrow \Delta(A)$ only depends on the current state input.

This work considers a neural network policy $\pi(s,a;\theta), \theta\in \Theta\subset\R^d$ as the state inputs are high-dimensional images. The RL problem for the stationary MDP (fixing $z$) is to find an optimal policy maximizing the expected cumulative rewards discounted by $\gamma\in (0,1]$:
\begin{align}
    \max_{\theta} J_z(\theta):=\E_{P(\cdot|\cdot;z), \pi(\cdot;\theta)}[\sum_{t=1}^H\gamma^t r(s_t,a_t)].
    \label{eq:rl}
\end{align}

We employ the policy gradient method to solve for the maximization using sample trajectories $\tau=(s_1,a_1, \ldots,s_H, a_H)$. The idea is to apply gradient descent with respect to the objective function $J_z(\theta)$. Following the policy gradient theorem \cite{sutton_PG}, we obtain $\nabla J_z(\theta)=\mathbb{E}[g(\tau;\theta)]$, where $ g(\tau;\theta)=\sum_{h=1}^H\nabla_\theta\log \pi(a_h|s_h;\theta)R^h(\tau)$,  and $R^h(\tau)=\sum_{t=h}^H r(s_t,a_t)$. In RL practice, the policy gradient $\nabla J_z(\theta)$ is replaced by its MC estimation since evaluating the exact value is intractable. Given a batch of trajectories $\mathcal{D}=\{\tau\}$ under the policy $\pi(\cdot|\cdot;\theta)$, the MC estimation is $\hat{\nabla} J(\theta,\mathcal{D}(\theta)):={1}/{|\mathcal{D}|}\sum_{\tau\in \mathcal{D}(\theta)} g(\tau;\theta)$. Our implementation applies the actor-critic (AC) method \cite{a3c}, a variant of the policy gradient, where the cumulative reward $R^h(\tau)$ in $g(\tau;\theta)$ is replaced with a $Q$-value estimator (critic) \cite{a3c}. The AC implementation is in Appendix \Cref{app:training}.

To improve the adaptability of RL policies, meta reinforcement learning (meta RL) aims to find a meta policy $\theta$ and an adaptation mapping $\Phi$ that returns satisfying rewards in a range of environments when the meta policy $\theta$ is updated using $\Phi$ within each environment. Mathematically, meta RL amounts to the following optimization problem \cite{finn18adapt_dy}:
\begin{align}
	\max_{\theta, \Phi} \quad  \E_{z\sim \rho_z}[J_z(\Phi(\theta,\mathcal{D}_z ))], \label{eq:meta_opt}
\end{align} 
where the initial distribution $\rho_z$ gives the probability that the agent is placed in the environment $z$. $D_z$ denotes a collection of sample trajectories under environment $z$ rolled out under $\theta$. The adaptation mapping $\Phi(\theta,\mathcal{D}_z)$ adapts the meta policy $\theta$ to a new policy $\theta_z$ fine-tuned for the specific environment $z$ using $\mathcal{D}_z$. For example, the adaptation mapping in MAML \cite{finn2017model} is defined by a policy-gradient update, i.e., $\Phi(\theta,\mathcal{D}_z )=\theta+\alpha\hat{\nabla}J_z(\theta, \mathcal{D}_z)$, where $\alpha$ is the step size to be optimized \cite{li2017meta}. Since the meta policy maximizes the average performance over a family of tasks, it provides a decent starting point for fine-tuning that requires far less data than training $\theta_z$ from scratch. As a result, the meta policy generalizes to a collection of tasks using sample trajectories.

\paragraph{Challenges in Online Adaptation} Meta RL formulation reviewed above does not fully address the limited adaptation issue in execution time since \eqref{eq:meta_opt} does not include the evolution of the hidden mode $z_{t+1}\sim p_z(\cdot|z_t)$. When interacting with a nonstationary environment online, the agent cannot collect enough trajectories $\mathcal{D}_{z_t}$ for adaptation as $z_t$ is time-varying. In addition, past experiences $\mathcal{D}_{z_t}$ only reveal information about the previous environments. The gradient updates based on past observations may not suffice to prepare the agent for the future. 

\section{Conjectural Online Lookahead Adaptation}
Instead of casting meta learning as a static optimization problem in \eqref{eq:meta_opt}, we consider learning the meta-adaptation online, where the agent updates its adaptation strategies at every step based on its observations. Unlike previous works \cite{finn18adapt_dy,rakelly19a,finn20_task_agnostic} aiming at learning a meta policy offline, our meta learning approach enables the agent to adapt to the changing environment. The following formally defines the online meta-adaptation learning (OMAL) problem.

Unlike \eqref{eq:meta_opt}, the online adaptation mapping relies on online observations $\cup_{k=1}^t\mathcal{I}_k$ instead of a batch of samples $\mathcal{D}_z$: the meta adaptation at time $t$ is defined as a mapping $\Phi_t(\theta):=\Phi(\theta,\cup_{k=1}^t\mathcal{I}_k) $. Let $r_t^\pi(s_t;\theta):=\E_{a\sim \pi(\cdot|s_t;\theta)}[r_t(s_t,a)]$ be the expected reward. The online meta-adaptation learning in the HM-MDP is given by 
\begin{align}
    \max_{\{\Phi_t\}_{t=1}^H}&\quad \E_{z_1,z_2,\cdots, z_H}[\sum_{t=1}^H r^\pi(s_t;\Phi_t(\theta))],\tag{OMAL}\label{eq:omal}
\end{align}
where $\quad z_{t+1}\sim p_z(\cdot|z_t), t=1,\ldots, H-1$, $\theta=\argmax \E_{z\sim \rho_z}[J_z(\theta)]$.
Some remarks on \eqref{eq:omal} are in order. First, similar to \eqref{eq:rl} and \eqref{eq:meta_opt}, \eqref{eq:omal} needs to be solved in a model-free manner as the transition kernel $P(\cdot|\cdot;z_t)$ regarding image inputs is too complicated to be modeled or estimated. Meanwhile, the hidden mode transition $p_z$ is also unknown to the agent. Second, the meta policy in \eqref{eq:omal} is obtained in a similar vein as in \eqref{eq:meta_opt}: $\theta$ maximizes the rewards across different environments, providing a decent base policy to be adapted later (see \Cref{algo:meta-policy} in Appendix). Unlike the offline operation in \eqref{eq:meta_opt} (i.e., finding $\theta$ and $\Phi$ before execution), \eqref{eq:omal} determines $\Phi_t$ on the fly in the execution phase based on $\cup_{k=1}^t\mathcal{I}_k$ to accommodate the nonstationary environment.    

Theoretically, searching for the optimal meta adaptation mappings $\{\Phi_t\}_{t=1}^H$ is equivalent to finding the optimal nonstationary policy $\{\theta_t\}_{t=1}^H$ in the HM-MDP, i.e., 
\begin{align}
    \max_{\{\theta_t\}_{t=1}^H} &\quad \E_{z_1,z_2,\cdots, z_H}[\sum_{t=1}^H r^\pi(s_t;\theta_t)].\label{eq:pomdp}
\end{align}
However, solving for \eqref{eq:pomdp} either requires domain knowledge regarding $P(\cdot|\cdot;z)$ and $p_z$ \cite{choi2000hidden,pineau11point,ross14onlinePOMDP} or black-box simulators producing successor states, observations, and rewards \cite{silver10mcpomdp}. When solving complex tasks such as the self-driving task considered in this work, these assumptions are no longer valid. The following presents a model-free real-time adaptation algorithm, where $\Phi_t$ is determined by a lookahead optimization conditional on the belief on the hidden mode.

 The meta policy in \eqref{eq:omal} is given by $\theta=\argmax\E_{z\sim \rho_z}[J_z(\theta)]$. Note that $z\in \mathcal{Z}$ in our case is a three-dimensional vector composed of three weather parameters. The hidden model space $\mathcal{Z}$ contains an infinite number of $z$, and it is impractical to train the policy over each environment. Similar to MAML \cite{finn2017model}, we sample a batch of modes $z$ using $\rho_z$ and train the meta policy over these sampled modes, as shown in \Cref{algo:meta-policy} in the Appendix. Note that the meta training program returns the stabilized policy iterate $\theta_k$, its corresponding mode sample batch $\widehat{\mathcal{Z}}$, and the associated sample trajectories $\{\widehat{\mathcal{D}}_z\}$ under mode $z$ and policy $\theta_k$. These outputs can be viewed as the agent's knowledge gained in the training phase and are later utilized in the online adaptation process.

\subsection{Lookahead Adaptation} In online execution, the agent forms a belief $b_t\in \Delta(\widehat{\mathcal{Z}})$ at time $t$ on the hidden mode based on its past observations $\cup_{k=1}^t \mathcal{I}_k$. Note that the support of the agent's belief is $\widehat{\mathcal{Z}}$ instead of the whole space $\mathcal{Z}$. The intuition is the agent always attributes the current weather pattern to a mixture of known patterns when facing unseen weather conditions. Then, the agent adapts to this new environment using its training experiences. Specifically, the agent conjectures that the environment would evolve according to $P(\cdot|\cdot;z)$ with probability $b_t(z)$ for a short horizon $K$. Given the agent's belief $b_t$ and its meta policy $\theta$, the trajectory of future $K$ steps  $\tau_t^{K}:=(s_t,a_t,\ldots,s_{t+K-1}, a_{t+K-1}, s_{t+K})$ follows $ q(\tau_t^K;b,\theta)$ defined as
\begin{align*}
   \prod_{k=0}^{K-1}\pi(a_{t+k}|s_{t+k};\theta) \prod_{k=0}^{K-1}[\sum_{z\in \mathcal{Z}}b_t(z)P(s_{t+k+1}|s_{t+k},a_{t+k}|z)].
\end{align*}

In order to maximize the forecast of the future performance in $K$ steps, the adapted policy $\theta_t=\Phi_t(\theta)$ should maximize the forecast future performance: $ \max_{\theta'\in \Theta}\E_{q(\tau_t^K;b,\theta')}\sum_{k=0}^{K-1} r(s_{t+k},a_{t+k})$. Note that the agent cannot access the distribution $q(\cdot;b,\theta')$ in the online setting and hence, can not use policy gradient methods to solve for the maximizer. Inspired by off-policy gradient methods \cite{queeney21off-ppo}, we approximate the solution to the future performance optimization using off-policy data $\{\widehat{\mathcal{D}}_z\}$.  

Following the approximation idea in trust region policy optimization (TRPO) \cite{schulman15trpo}, the policy search over a lookahead horizon under the current conjecture, referred to as conjectural lookahead optimization, can be reformulated as  
\begin{align}\label{eq:trpo}
		\max_{\theta'\in\Theta} & \ \E_{q(\cdot;b_t,\theta)}\left[\prod_{k=0}^{K-1}\frac{\pi(a_{t+k}|s_{t+k};\theta')}{\pi(a_{t+k}|s_{t+k};\theta)}\sum_{k=0}^{K-1} r(s_{t+k},a_{t+k})\right]\tag{CLO}\\
	\st &\quad \E_{s\sim q} D_{KL}(\pi(\cdot|s;\theta), \pi(\cdot|s;\theta'))\leq \delta,\nonumber
\end{align}
where $D_{KL}$ is the Kullback-Leibler divergence. In the KL divergence constraint, we slightly abuse the notation $q(\cdot)$ to denote the discounted state visiting frequency $s\sim q$. 

Note that the objective function in \eqref{eq:trpo} is equivalent to the future performance under $\theta'$: $ \E_{q(\tau_t^K;b,\theta')}\sum_{k=0}^{K-1} r(s_{t+k},a_{t+k})$. This is because the distribution shift between $q(\tau_t^K;b,\theta')$ and $q(\tau_t^K;b,\theta)$ in \eqref{eq:trpo} is compensated by the ratio $\prod_{k=0}^{K-1}\frac{\pi(a_{t+k}|s_{t+k};\theta')}{\pi(a_{t+k}|s_{t+k};\theta)}$.  The intuition is that the expectation in \eqref{eq:trpo} can be approximated using $\mathcal{D}_z(\theta)$ collected in the meta training. When $\theta'$ is close to the base policy $\theta$ in terms of KL divergence, the estimated objective in \eqref{eq:trpo} using sample trajectories under $\theta$ returns a good approximation to $\E_{q(\tau^K;b,\theta')}\sum_{k=0}^{K-1} r(s_{t+k},a_{t+k})$. This approximation is detailed in the following subsection.

\subsection{Off-policy Approximation}
We first simplify \eqref{eq:trpo} by linearizing the objective and approximating the constraints using a quadratic surrogate as introduced in \cite{schulman15trpo,achiam17cpo}. Denote by $L(\theta';\theta)$ the objective function in \eqref{eq:trpo}. A first-order linearization of the objective at $\theta'=\theta$ is given by $L(\theta';\theta)\approx \nabla_{\theta'}L(\theta';\theta)|_{\theta'=\theta}(\theta'-\theta)$. Note that $\theta'$ is the decision variable, and $\theta$ is the known meta policy. The gradient $\nabla_{\theta'}L(\theta';\theta)|_{\theta'=\theta}$ is exactly the policy gradient under the trajectory distribution $q(\cdot;b_t,\theta)$. Using the notions introduced in \Cref{sec:model}, we obtain $\nabla_{\theta'}L(\theta';\theta)|_{\theta'=\theta}=\E_{q(\cdot;b_t,\theta)}[g(\tau_t^K;\theta)]=\sum_{z\in\widehat{\mathcal{Z}}} b(z)\E_{q(\cdot;z,\theta)} [g(\tau_t^K;\theta)]$. Note that $\tau_t^K$ is the future $K$-step trajectory, which is not available to the agent at time $t$, and a substitute is its counterpart in $\widehat{D}_z$: the sample trajectory within the same time frame $[t,t+K]$ in meta training. Denote this counterpart by $\hat{\tau}_t^K$, then $\E_{q(\cdot;b_t,\theta)}[g(\tau_t^K;\theta)]\approx \sum_{z\in\widehat{\mathcal{Z}}} b(z) g(\hat{\tau}_t^K;\theta)$, and we denote its sample estimate by $\hat{g}(b;\theta)$. The distribution $q(\cdot|z,\theta)$ is defined similarly as $q(\cdot|b,\theta)$. More details on this distribution and $\hat{g}(b;\theta)$ are in Appendix \Cref{app:opa}.

We consider a second-order approximate for the KL-divergence constraint because the gradient of $D_{KL}$ at $\theta$ is zero. In this case, the approximated constraint is $\frac{1}{2}(\theta-\theta')^\tp A(\theta-\theta')\leq \delta$, where $A_{ij}=\frac{\partial}{\partial\theta_i\partial\theta_j}\E_{s\sim q} D_{KL}(\pi(\cdot|s;\theta), \pi(\cdot|s;\theta'))|_{\theta'=\theta}$. Let $\hat{A}$ be the sample estimate using $\widehat{\tau}_t^K$ (see Appendix \Cref{app:opa}), then the off-policy approximate to \eqref{eq:trpo} is given by 
\begin{align}
\label{eq:opa}
	\max_{\theta'}&\quad\hat{g}(b;\theta)(\theta'-\theta)\\
	\st &\quad \frac{1}{2}(\theta-\theta')^\tp \hat{A}(\theta-\theta')\leq \delta.\nonumber
\end{align}
To find such $\theta'$, we compute the search direction $d\theta=\hat{A}^{-1}\hat{g}$ efficiently using conjugate gradient method as in original TRPO implementation \cite{schulman15trpo}. Generally, the step size is determined by backtracking line search, yet we find that fixed step size also achieves impressive results in experiments.  
\subsection{Belief Calibration} 
\label{sec:belief-calibration}
The last piece in our online adaptation learning is belief calibration: how the agent infers the currently hidden mode $z_t$ based on past experiences and updates its belief based on new observations. A general-purpose strategy is to train an inference network through a variational inference approach \cite{rakelly19a}. Considering the significant visual differences caused by weather conditions, we adopt a more straightforward approach based on image classification and Bayesian filtering. To simplify the exposition, we only consider two kinds of weather: cloudy (denoted the mode by $z_c$) and rainy (denoted by $z_r$) in meta training $\widehat{Z}=\{z_r, z_c\}$.  

\paragraph{Image Classification}The image classifier is a binary classifier based on ResNet architecture \cite{resnet}, which is trained using cross-entropy loss. The data sets (training, validation, and testing) include RGB camera images generated by CARLA. The input is the current camera image, and the output is the probability of that image being of the rainy type. The training details are included in Appendix \Cref{app:training}.

Given an arbitrary image, the underlying true label is created by measuring its corresponding parameter distance to the default weather setup in the CARLA simulator (see Appendix \Cref{app:training}). Finally, we remark that the true labels and the hidden mode $z_r, z_c$ are revealed to the agent in the training phase. Yet, in the online execution, only online observations $\mathcal{I}_t=\{s_t,a_{t-1},r_{t-1}\}$ are available.
\paragraph{Bayesian Filtering} Denote the trained classifier by $f:\mathcal{S}\rightarrow \Delta(\widehat{\mathcal{Z}})$, where $f(z_r;s)$ denotes the output probability of $s$ being an image input under the mode $z_r$. During the online execution phase, given an state input $s_t$ and the previous belief $b_{t-1}\in \Delta(\widehat{Z})$, the updated belief is 
\begin{align}
	b_{t}(z_r)=\frac{b_{t-1}(z_r)f(z_r;s_t)}{\sum_{z\in \widehat{Z}}b_{t-1}(z)f(z;s_t)}.\label{eq:bf}
\end{align}  
The intuition behind \eqref{eq:bf} is that it better captures the temporal correlation among image sequences than the vanilla classifier $f$. Note that this continuously changing weather is realized through varying three weather parameters. Hence, two adjacent images,  as shown in \Cref{fig:image_input}, display a temporal correlation: what follows a rain image is highly likely to be another rain image. Considering this correlation, we use Bayesian filtering,
where the prior reveals information about previous images. As shown in an ablation study in \Cref{app:ablation}, \eqref{eq:bf} does increase the belief accuracy (to be defined later in \Cref{app:ablation}) and associated rewards. Finally, we conclude this section with the pseudocode of the proposed conjectural online lookahead adaptation (COLA) algorithm presented in \Cref{algo:cola}. 
\begin{algorithm}	
	\caption{\textbf{C}onjectural \textbf{O}nline \textbf{L}ookahead \textbf{A}daptation}\label{algo:cola}
		\begin{algorithmic}
			\State  \textbf{Input} The meta policy $\theta$, classifier $f$, training samples $\{\mathcal{D}_z\}_{z\in \widehat{Z}}$, sample batch size $M$, lookahead horizon length $K$, and  step size $\alpha$. Set $\theta_1=\theta$.
			\For {$t\in \{1,2,\ldots, H\}$}
			\State Obtain the image input $s_t$ from the camera;
			\State Implement the action using $\pi(\cdot|s_t;\theta_t)$;
			\State Obtain the probability $f(z;s_t)$ from the classifier;
			\State Update the belief using \eqref{eq:bf}; 
			\State Sample $M$ trajectories $\hat{\tau}_t^K$ under  $\hat{z}$ from $\{\mathcal{D}_z\}_{z\in \widehat{Z}}$;
			\State Obtain $\theta'$ by solving \eqref{eq:opa}, and let $\theta_{t+1}=\theta'$;
			\EndFor
		\end{algorithmic}
	\end{algorithm}\vspace{-6mm}
\vspace{-0.5mm}
\section{Experiments}
\label{sec:exp}
This section evaluates the proposed COLA algorithm using CARLA simulator \cite{carla}. The HM-MDP setup for the lane-keeping task (i.e., hidden mode transition, the reward function) is presented in Appendix \Cref{app:setup}. This section reports experimental results under $K=10$, and experiments under other choices are included in Appendix \Cref{app:tradeoff}. 

We compare COLA with the following RL baselines. 1) The base policy $\theta_{base}$: a stationary policy trained under dynamic weather conditions, i.e., $\theta_{base}=\argmax \E_{z_1,z_2,\cdots, z_H}[\sum_{t=1}^H r^\pi(s_t;\theta)]$. 2) The MAML policy $\theta_{MAML}$: the solution to \eqref{eq:meta_opt} with $\Phi$ being the policy gradient computed using the \emph{past} $K$-step trajectory. 3) The optimal policy under the mixture of $Q$ functions $\theta_{Q_{mix}}$: the action at each time step is determined by $a_t=\argmax_{a\in \mathcal{A}}\sum_{z\in \widehat{Z}} b_t(z)Q_z(s_t,a)$, where $Q_z$ is the $Q$ function trained under the stationary MDP with fixed mode $z$. In our experiment, this mixture $\sum_{z\in \widehat{Z}} b_t(z)Q_z(s_t,a)$ is composed of two $Q$ functions $Q_c, Q_r$ (sharing the same network structure as the critic) under the cloudy and rainy conditions, respectively. 4) The oracle policy $\theta_{oracle}$: an approximate solution to \eqref{eq:pomdp},  obtained by solving \eqref{eq:trpo} using authentic future trajectories generated by the simulator.          

\begin{figure}[h]
	\begin{subfigure}{0.23\textwidth}
	\centering
	\includegraphics[width=0.9\textwidth]{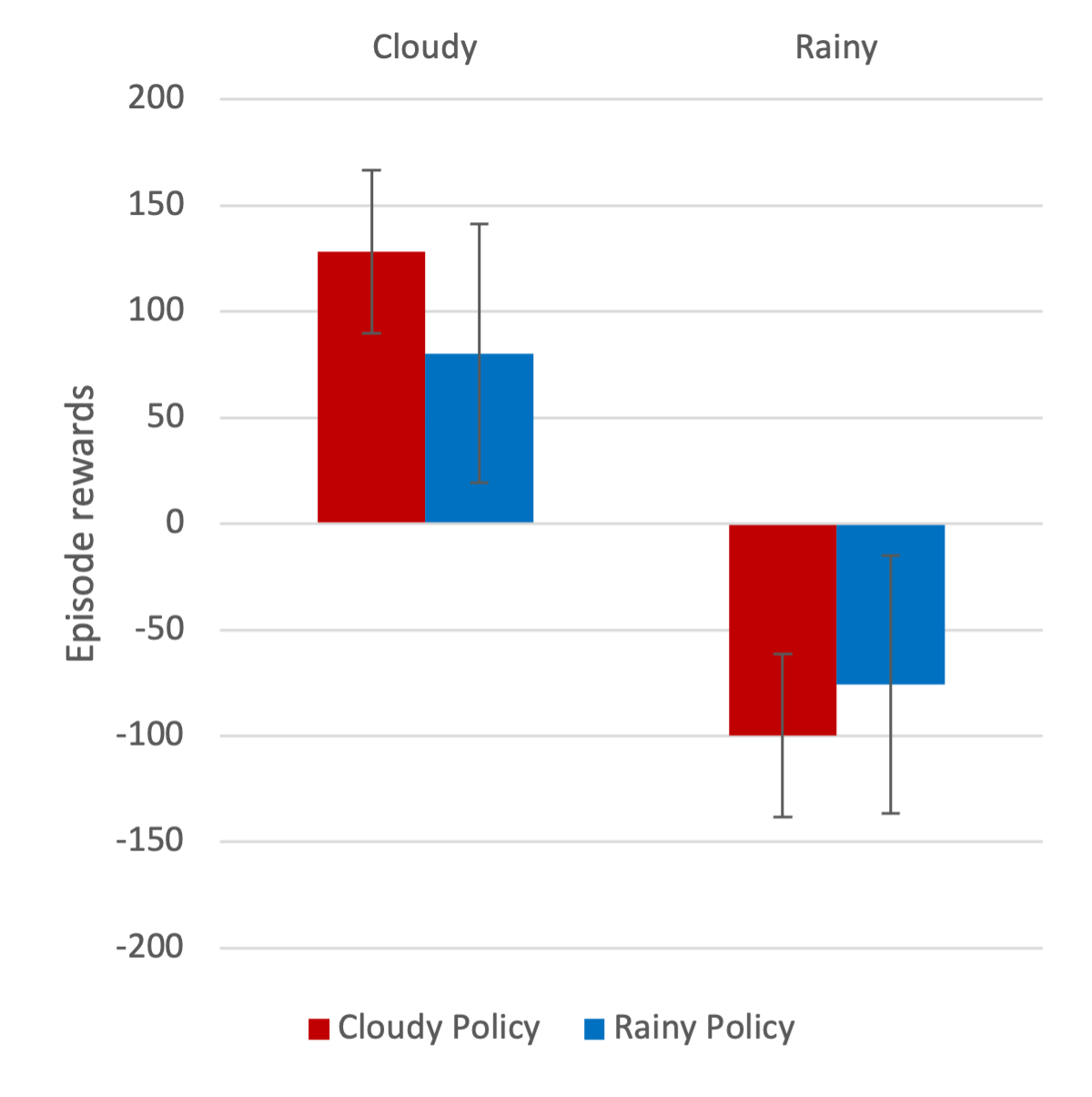}	
	\caption{}
	\label{fig:poor}
	\end{subfigure}
	\begin{subfigure}{0.23\textwidth}
		\centering
		\includegraphics[width=0.9\textwidth]{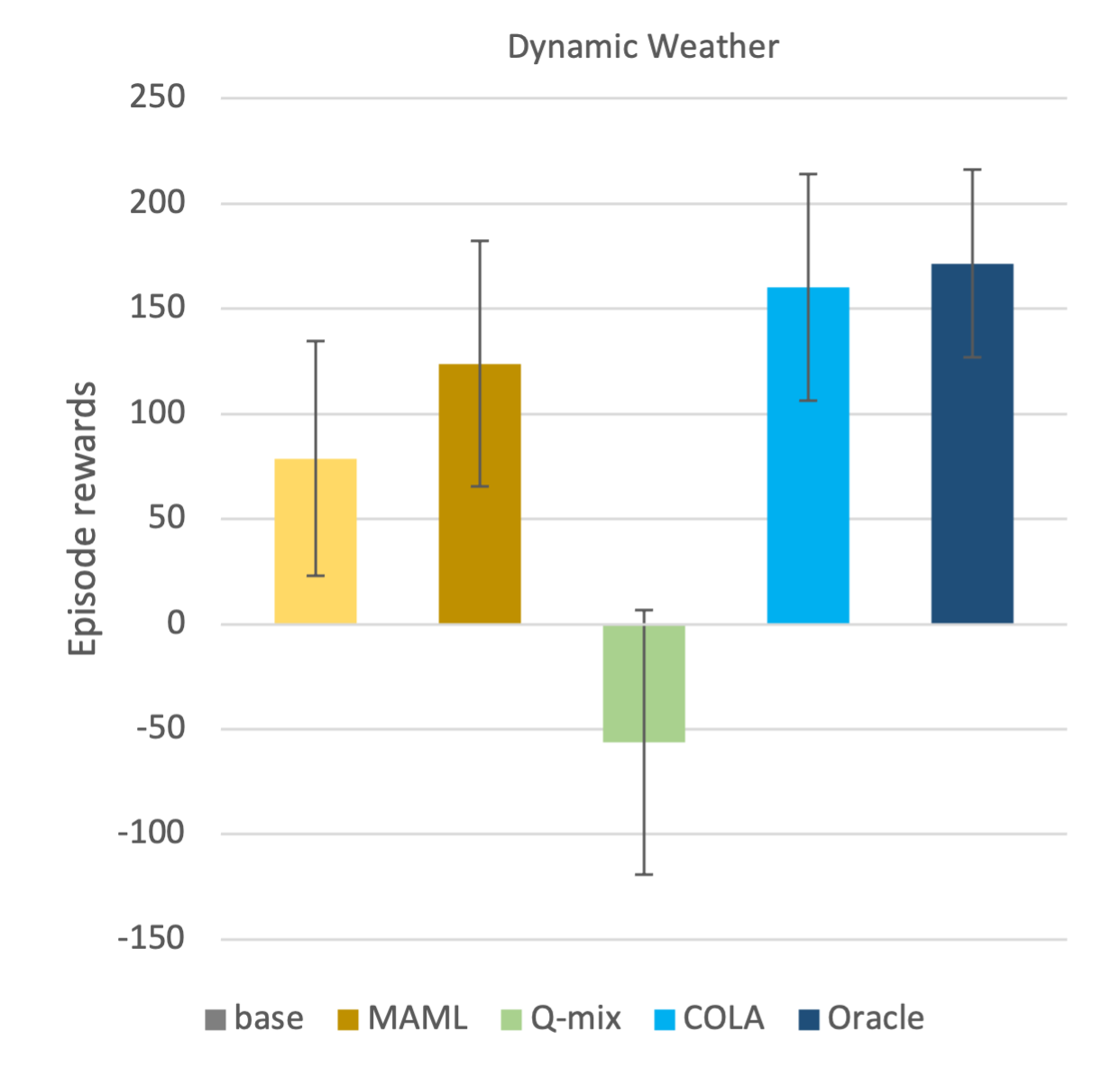}
		\caption{}
		\label{fig:compare}
	\end{subfigure}
	\begin{subfigure}{0.23\textwidth}
		\centering
		\includegraphics[width=0.9\textwidth]{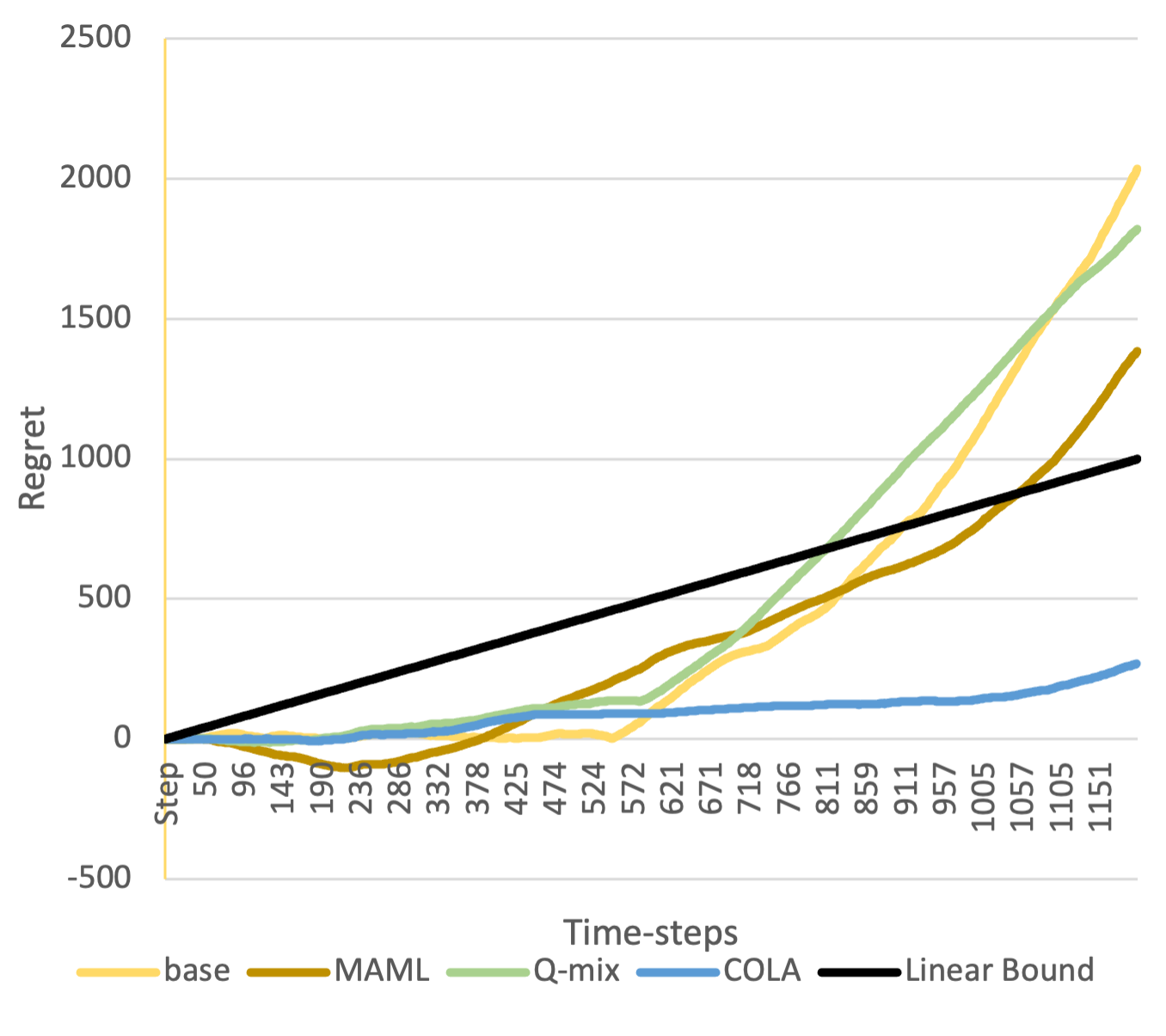}
		\label{fig:regret}
		\caption{}
	\end{subfigure}
		\begin{subfigure}{0.23\textwidth}
	    \centering
	    \includegraphics[width=0.9\textwidth]{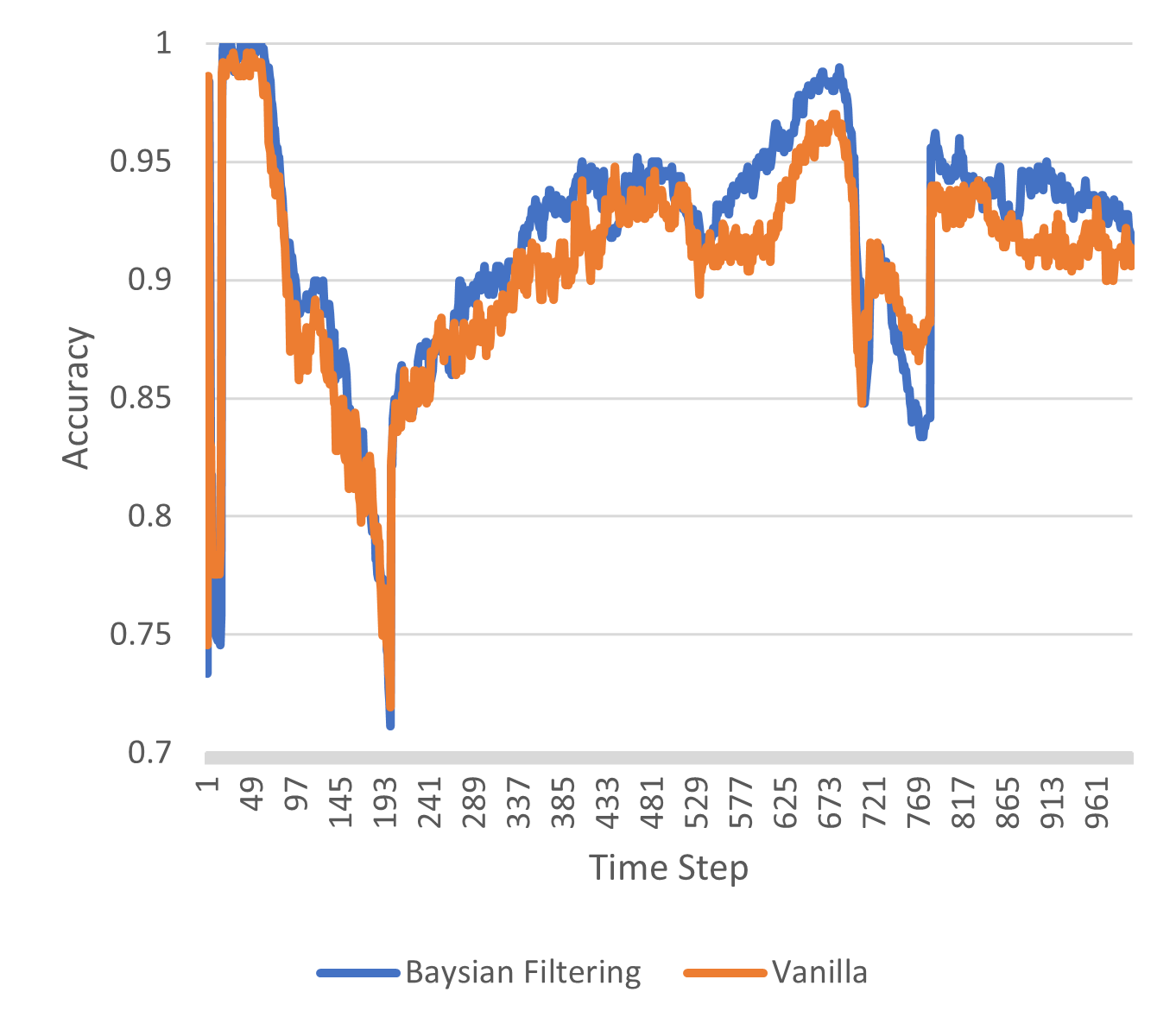}
	    \caption{}
	    \label{fig:transient}
	\end{subfigure}
	\caption{Evaluations of COLA algorithms in the lane-keeping task under nonstationary weather conditions (averaged over 500 episodes under 10 random seeds). (a) The average rewards of the cloudy-trained the rainy-trained policies in cloudy and rainy conditions. The policy trained under one environment does not generalize to the other. (b) The average rewards of the COLA policy and the four baseline policies. COLA achieves comparable results as the oracle policy and outperforms the other baselines. (c) The regret grows under different policies. Only COLA policy achieves sub-linear regret growth [below the linear bound $O(T)$]. (d) The evolution of transient accuracies (averaged over 500 episodes) with respect to time step $t$ under Bayesian filtering and vanilla classification in dynamic weather.} 
	\label{fig:exp}\vspace{-6mm}
\end{figure}

The experimental results are summarized in \Cref{fig:exp}. As shown in \Cref{fig:compare}, the COLA policy outperforms $\theta_{base}$, $\theta_{MAML}$, and $\theta_{Q_{mix}}$, showing that the lookahead optimization \eqref{eq:trpo} better prepares the agent for the future environment changes. Note that $\theta_{Q_{mix}}$ returns the worst outcome in our experiment, suggesting that the naive adaptation by averaging $Q$ function does not work: the nonstationary environment given by HM-MDP does not equal an average of multiple stationary MDPs. The agent needs to consider the temporal correlation as discussed in \Cref{sec:belief-calibration}.  

To evaluate the effectiveness of COLA-based online adaptation,  we follow the regret notion used in online learning \cite{shai_online} and consider the performance differences between COLA (the other three baselines) and the oracle policy. The performance difference or regret up to time $T$ is defined as $\operatorname{Reg}(T)=\E[\sum_{t=1}^T r^\pi(s_t;\theta_{oracle})-\sum_{t=1}^T r^{\pi}(s_t;\Phi_t(\theta))]$. If the regret grows sublinearly, i.e., $\operatorname{Reg}(T)< O(T)$, then the corresponding policy achieves the same performance as the oracle asymptotically. As shown in \Cref{fig:regret}, the proposed COLA is the only algorithm that achieves comparable results as the oracle policy.   
 \vspace{-0mm}

\subsection{Knowledge Transfer through Off-policy Gradients}
\begin{figure}
	\begin{subfigure}{0.24\textwidth}
		\centering
		\includegraphics[width=1\textwidth]{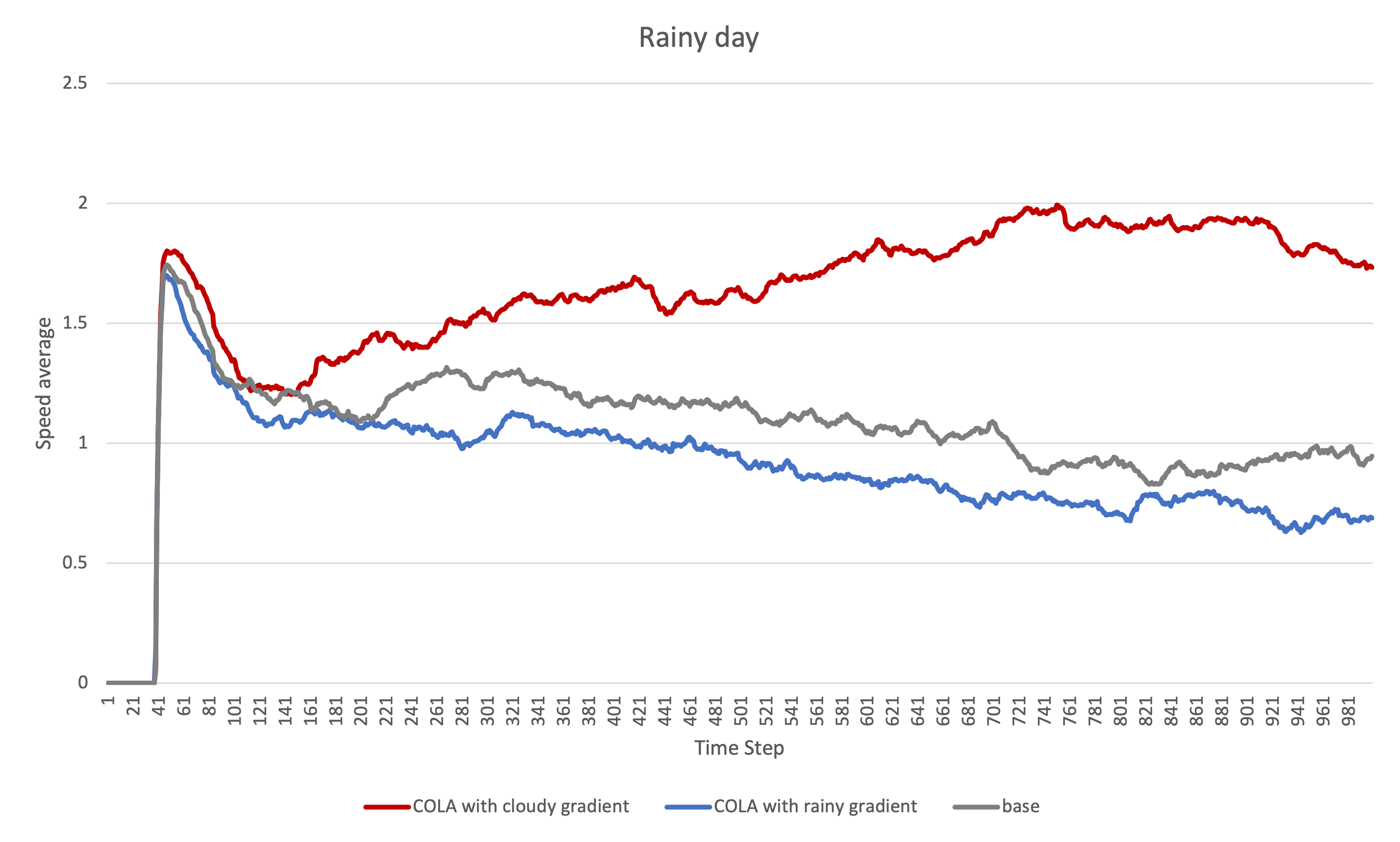}
		\caption{}\label{fig:rain-speed}
	\end{subfigure}
		\begin{subfigure}{0.24\textwidth}
		\centering
		\includegraphics[width=1\textwidth]{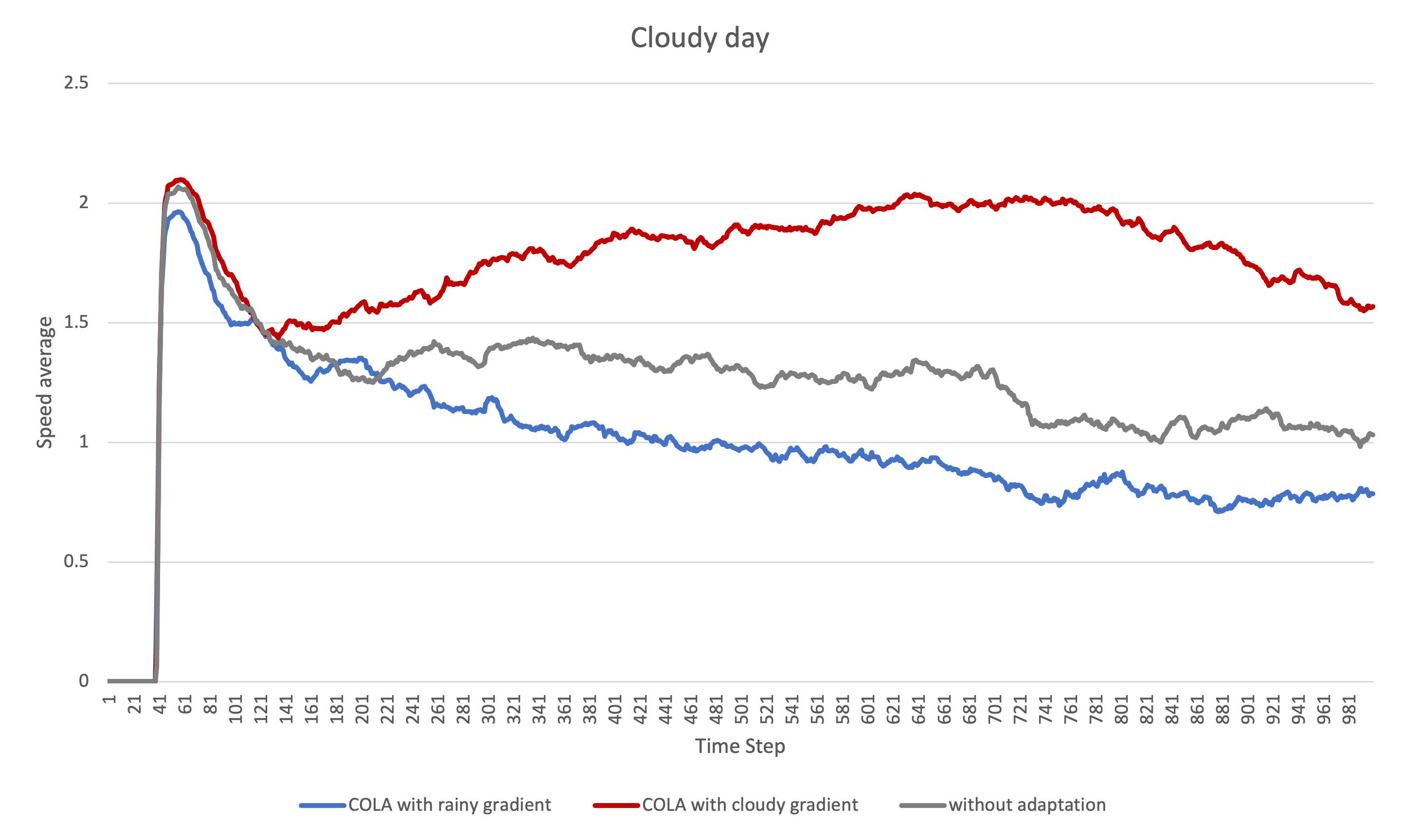}
		\caption{}\label{fig:cloudy-speed}
	\end{subfigure}
		\caption{Figure (a) [or (b)] presents average speeds at each time step within the horizon (average of 500 episodes) with $\hat{g}_c$ (red curves) and $\hat{g}_r$ (blue curves) being used to adapt the meta policy (grey curves) under the cloudy [or rainy in (b)] mode. Using $\hat{g}_c$ [or $\hat{g}_r$] always leads to aggressive [or conservative] driving, regardless of the environment. }\vspace{-6mm}
\end{figure}

 This subsection presents an empirical explanation of COLA's adaptability: the knowledge regarding the speed control learned in meta training is carried over to online execution through the off-policy gradient $\hat{g}$ in \eqref{eq:opa}. As observed in \Cref{tab:speed} in Appendix, when driving in a rainy environment (i.e., $z_r$),  the agent tends to drive slowly to avoid crossing the yellow line, as the visibility is limited. In contrast, it increases the cruise speed when the rain stops (i.e., $z_c$). Denote the off-policy gradients under $z_r, z_c$ by $\hat{g}_r$  and $\hat{g}_c$, respectively. Then, solving \eqref{eq:opa} using $\hat{g}_r$ ($\hat{g}_c$) leads to a conservative (aggressive) driving in terms of speeds as shown in \Cref{fig:rain-speed} (\Cref{fig:cloudy-speed}), regardless of the environment changes. This result suggests that the knowledge gained in meta training is encoded into the off-policy gradients and is later retrieved according to the belief. Hence, the success of COLA relies on a decent belief calibration, where the belief accurately reveals the underlying mode. \Cref{app:ablation} justifies our choice of Bayesian filtering.  

\subsection{An Ablation Study on Bayesian Filtering}
\label{app:ablation}
For simplicity, we use the 0-1 loss to characterize the misclassification in the online implementation in this ablation study. Denote by $\hat{i}^\nu_t\in \{0, 1\}$ the label returned by the classifier (i.e., the class assigned with a higher probability; 1 denotes the rain class) at time $t\in \{1,\ldots, H\}$ in the $\nu$-th episode (500 episodes in total).  Then the loss at time $t$ is $\mathbf{1}_{\{\hat{i}^\nu_t\neq i^\nu_t\}}$, where $i^\nu_t$ is the true label. We define the transient accuracy at time $t$ as $
	\text{Transient Accuracy}=\frac{\sum_{\nu=1}^{500} 1- \mathbf{1}_{\{\hat{i}^\nu_t\neq i^\nu_t\}}}{500}=\frac{\sum_{\nu=1}^{500}  \mathbf{1}_{\{\hat{i}^\nu_t=i^\nu_t\}}}{500},
$    
which reflects the accuracy of the vanilla classifier at time $t$. The average transient accuracy over the whole episode horizon $H$ is defined as the episodic accuracy, i.e., $
	\text{Episodic Accuracy}=\frac{\sum_{\nu=1}^{500}\sum_{t=1}^H  \mathbf{1}_{\{\hat{i}^\nu_t=i^\nu_t\}}}{500 H}$.
These accuracy metrics can also be extended to the Bayesian filter, where the outcome label corresponds to the class with a higher probability.

We compare the transient and episodic accuracy as well as the average rewards under Bayesian filtering and vanilla classification, and the results are summarized in \Cref{tab:episodic} and \Cref{fig:transient}. Even though the two methods' episodic accuracy under two methods differ litter, as shown in \Cref{tab:episodic}, the difference regarding the average rewards indicates that Bayesian filtering is better suited to handle nonstationary weather conditions. As shown in \Cref{fig:transient}, Bayesian filtering outperforms the classifier for most of the horizon in terms of transient accuracy. The reason behind its success is that Bayesian filtering better captures the temporal correlation among the image sequence and returns more accurate predictions than the vanilla classifier at every step, as the prior distribution is incorporated into the current belief. It is anticipated that higher transient accuracy leads to higher mean rewards since the lookahead optimization [see \eqref{eq:trpo}] depends on the belief at every time step.
\begin{table}
	\centering
	\begin{tabular}{ccc}
	\toprule
		& Bayesian Filtering & Vanilla Classification\\
		\midrule
		Episodic Accuracy &  91.18\% & 90.41\%\\
		Average Reward & $152.37\pm 49.58$ & $122.44\pm54.80$ \\
		\bottomrule
	\end{tabular}
	\caption{Comparisons of episodic accuracies and average rewards under dynamic weather.}
	\label{tab:episodic}\vspace{-6mm}
\end{table}

\section{Related Works}
\paragraph{Reinforcement Learning for Self Driving} In most of the self-driving tasks studied in the literature \cite{schwarting18ad_review}, such as lane-keeping and changing \cite{Sallab2016EndtoEndDR,Wang2018ARL}, overtaking \cite{ngai11overtaking}, and negotiating intersections \cite{chen21end2end}, the proposed RL algorithms are developed in an offline approach. The agent first learns a proficient policy tailored to the specific problem in advance. Then the learned policy is tested in the same environment as the training one without any adaptations. Instead of focusing on a specific RL task (stationary MDP) as in the above works, the proposed OMAL tackles a nonstationary environment modeled as an HM-MDP. Closely related to our work, a MAML-based approach is proposed in  \cite{jaafra19context-meta} to improve the generalization of RL policies when driving in diverse environments. However, the adaptation strategy in \cite{jaafra19context-meta} is obtained by offline training and only tested in stationary environments. In contrast, our work investigates online adaptation in a nonstationary environment. The proposed COLA enables the agent to adapt rapidly to the changing environment in an online manner by solving \eqref{eq:trpo} approximately in real time. 

\paragraph{Online Meta Learning} To handle nonstationary environments or a streaming series of  different tasks, recent efforts on meta learning have integrated online learning ideas with meta learning, leading to a new paradigm called online meta learning  \cite{finn19ftml,finn20_task_agnostic} or continual meta learning \cite{xie20lifelong, osaka}. Similar to our nonstationary environment setup, \cite{finn20_task_agnostic,osaka,xie20lifelong} focuses on continual learning under a nonstationary data sequence involving different tasks, where the nonstationarity is governed by latent variables. Despite their differences in latent variable estimate, these works utilize on-policy gradient adaptation: the agent needs to collect $\mathcal{D}_z$ and then adapt $\theta_z=\Phi(\theta,\mathcal{D}_z)$ [see \eqref{eq:meta_opt}].  Note that on-policy gradient adaptation requires a few sample data under the current task, which may not be suitable for online control. The agent can only collect one sample trajectory online, incurring a significant variance. In addition, by the time the policy is updated using the old samples, the environment has already changed, rendering the adapted policy no longer helpful. In contrast, COLA better prepares the agent for the incoming task/environment by maximizing the forecast of the future performance [see \eqref{eq:trpo}]. Moreover, the adaptation in COLA can be performed in real time using off-policy data.
\vspace{-0.5em}
\section{Conclusion}
This work proposes an online meta reinforcement learning algorithm based on conjectural online lookahead adaptation (COLA) that adapts the policy to the nonstationary environment modeled as a hidden-mode MDP. COLA enables prompt online adaptation by solving the conjectural lookahead optimization \eqref{eq:trpo} on the fly using off-policy data, where the agent's conjecture (belief) on the hidden mode is updated in a Bayesian manner. 


\bibliographystyle{IEEEtran}
\bibliography{cola}

\clearpage

\begin{appendices}
\renewcommand{\theequation}{\thesection.\arabic{equation}}
\section{Meta Training}
The meta training procedure is described in \Cref{algo:meta-policy}.
\begin{algorithm}	
\caption{Meta Training}\label{algo:meta-policy}
	\begin{algorithmic}
		\State  \textbf{Input} Initialization $\theta_0$, $k=0$, step size $\alpha$.
		\While{not converge}
		\State Draw the $k$-th batch of modes (weather conditions) $\mathcal{Z}^k_{train}=\{z\}, z\in \mathcal{Z}$;
		\For{\text{each environment} $z$ }
		\State Sample a batch of trajectories $\mathcal{D}_z(\theta_k)$ under $P(\cdot|\cdot;z)$ and $\pi(\cdot|\theta_k)$;
		\State Compute the policy gradient $\hat{\nabla}J_z(\theta_k,\mathcal{D}_z)$
		\EndFor
		\State $\theta_{k+1}=\theta_k+\alpha/|\mathcal{Z}^k_{train}| \sum_{z} \hat{\nabla}J_z(\theta_k,\mathcal{D}_z)$;
		\State $k=k+1$.
		\EndWhile
		\State\textbf{Return } $\theta_{k}$, $\widehat{\mathcal{Z}}=\mathcal{Z}^k_{train}$, $\{\widehat{\mathcal{D}}_z\}=\{\mathcal{D}_z(\theta_k)\}_{z\in \mathcal{Z}^k_{train}}$. 
	\end{algorithmic}
\end{algorithm}
Our experiment applies two default weather conditions in CARLA world: the cloudy and rainy in meta training. In this case, $\widehat{\mathcal{Z}}=\{z_r, z_c\}$ and $\widehat{\mathcal{D}}_{z_r}$ corresponds to the sample trajectories collected under the rainy condition. Likewise, $\widehat{\mathcal{D}}_{z_c}$ denotes those under the cloudy one. 

\section{Off-policy Approximation}
\label{app:opa}
To reduce the variance of the estimate $\hat{g}(b,\theta)$, we use the sample mean of multiple trajectories, i.e., 
\begin{align*}
	&\E_{q(\cdot;b_t,\theta)}[g(\tau_t^K;\theta)]\approx  \sum_{z\in\widehat{\mathcal{Z}}} \sum_{\hat{\tau}_t^K\in \widehat{\mathcal{D}}_z}1/|\widehat{\mathcal{D}_z}|b(z) g(\hat{\tau}_t^K;\theta),\\
	&q(\tau_t^K;z,\theta)=\prod_{k=0}^{K-1}\pi(a_{t+k}|s_{t+k};\theta)P(s_{t+k+1}|s_{t+k},a_{t+k}|z).
\end{align*}

The sample estimate of $A$ using the off-policy data is $$\hat{A}_{ij}=\frac{1}{K}\sum_{k=0}^{K-1}\frac{\partial}{\partial\theta_i\partial\theta_j}D_{KL}(\pi(\cdot|s_{t+k};\theta), \pi(\cdot|s_{t+k};\theta')).$$ One can also take the sample mean of $\hat{A}$ over multiple trajectories $\hat{\tau}_t^K$.

\begin{figure}
	\centering
	\includegraphics[width=0.5\textwidth]{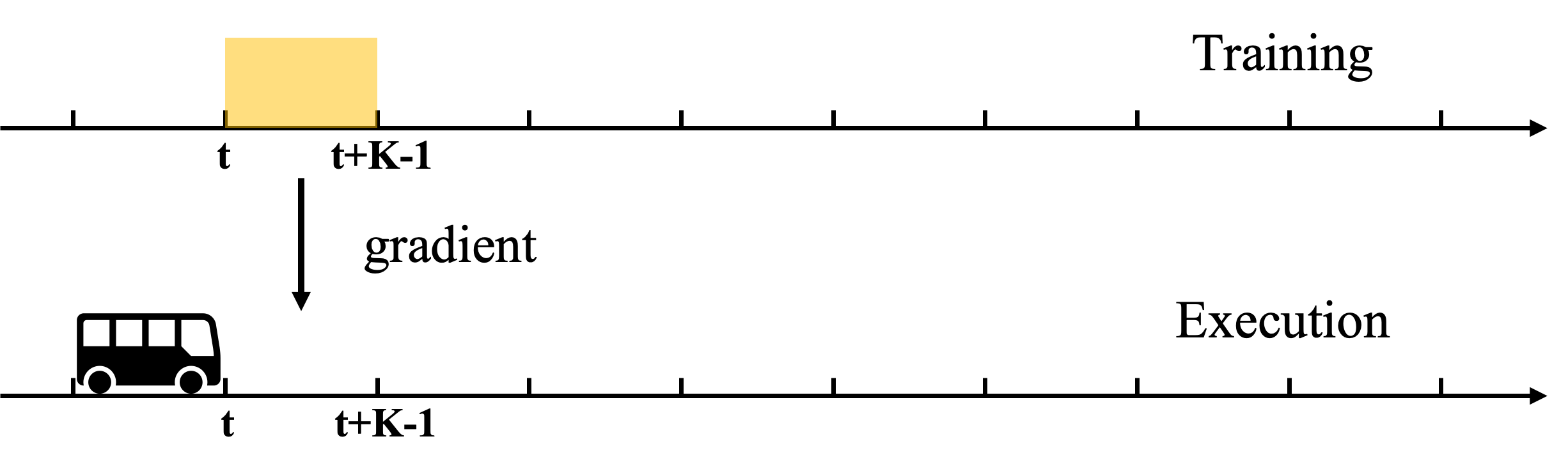}
	\caption{Off-policy Approximation. The future $K$-step trajectory $\tau_t^K$ is not available at time $t$ in online execution. A substitute is its counterpart in $\widehat{D}_z$: the sample trajectory $\hat{\tau}_t^K$ within the same time frame $[t,t+K]$ in meta training. }
	\label{fig:off-tra}
\end{figure}

It should be noted that $\hat{\tau}_t^K$ is not an i.i.d copy of $\tau_t^K$, as the initial states $s_t\in \tau_t^K$ and $\hat{s}_t\in \hat{\tau}_t^K$ are subject to different distributions determined by previous policies $\{\theta_k\}_{k=1}^t$. In other words, as the online adaptation updates $\theta_t$ at each time step by solving \eqref{eq:trpo}, $\theta_t$ deviates from the meta policy $\theta$, leading to a distribution shift between $\tau_t^K$ and $\hat{\tau}_t^K$. Hence, $\hat{\tau}_t^K$ is referred to as the off-policy (i.e., not following the same policy) data with respect to $\tau_t^K$. Thanks to the KL-constraint in \eqref{eq:trpo}, such deviation is manageable through the constraint $\delta$, and a mechanism has been proposed in \cite{queeney21off-ppo} to address the distribution shift. Finally, \Cref{fig:off-tra} provides a schematic illustration on how the training data is collevcted and reused in the online implementation, and \Cref{tab:speed}  summarizes the average speeds under three different weather setups. 

\begin{table}[H]

    \centering
    \begin{tabular}{cccc}
    \toprule
    Algorithm & Cloudy & Rainy & Dynamic\\
    \midrule
     {COLA}   &1.26 &1.07 &1.53\\
    \bottomrule
    \end{tabular}
    \caption{Average speeds of 500 episodes under COLA in three weather setups. }
    \label{tab:speed}
\end{table}
\section{Experiement Setup}\label{app:setup}
Our experiments use CARLA-0.9.4 \cite{carla} as the autonomous driving simulator. On top of the CARLA, we modify the API: Multi-Agent Connected Autonomous Driving (MACAD) Gym ~\cite{macad-gym} to facilitate communications between learning algorithms and environments. To simulate various traffic conditions,  we enable non-player vehicles controlled by the CARLA server. 

Three kinds of weather conditions are considered in experiments: stationary cloudy, rainy, and dynamic weather conditions. For stationary weather conditions, we use two pre-defined weather setup: \textit{CloudyNoon} and \textit{MidRainyNoon} provided by CARLA.  For the dynamic environment, the changing weather in  experiments is realized by varying three weather parameters: cloudiness, precipitation, and puddles. The changing pattern of dynamic weather is controlled by the parameters update frequency, the magnitude of each update, and the initial values. The basic idea is that as the clouds get thick and dense, the rain starts to fall, and the ground gets wet; when the rain stops, the clouds get thin, and the ground turns dry.

\paragraph{States, Actions, and Rewards} The policy model uses the current and previous 3-channel RGB images from the front-facing camera (attached to the car hood) as the state input. Basic vehicle controls in CARLA include throttle, steer, brake, hand brake, and reverse.  Our experiments only consider the first three, and the agent employs discrete actions by providing throttle, brake, and steer values. Specifically, these values are organized as two-dimensional vectors [throttle/brake, steer]: the first entry indicates the throttle or brake values, while the other represents the steer ones. In summary, the action space is a 2-dimensional discrete set, and its elements are shown in \Cref{tab:action_space}. 

\begin{table}[h]
	\centering
	\caption{Discrete control actions in the experiments}
	\label{tab:action_space}
	\begin{tabular}{ll}	
	\toprule 
	Action Vector & Control \\
	\midrule
	$[0.0, 0.0]$& Coast  \\
	$[0.0, -0.5]$ &Turn Left\\
	$[0.0, 0.5]$ &Turn Right\\
	$[1.0, 0.0]$ &Forward\\
	$[-0.5, 0.0]$ &Brake\\
	$[0.5, -0.5]$ &Forward Left\\
	$[0.5, 0.5]$ &Forward Right\\
	$[-0.5, -0.5]$ &Brake Left\\
	$[-0.5, 0.5]$ &Brake Right\\
	$[0.5, 0.0]$ &Bear Forward\\
	$[0.75, 1.0]$ &Sharp Right\\
	$[0.75, -1.0]$ &Sharp Left\\
	$[-0.5, 1.0]$ &Brake Sharp Right\\
	$[-0.5, -1.0]$ &Brake Sharp Left\\
	$[0.5, 0.75]$ &Accelerate Forward Inferior Sharp Right\\
	$[0.5, -0.75]$ &Accelerate Forward Inferior Sharp Left\\
	$[0.75, 0.0]$ &Accelerate Forward\\
	$[-1.0, 0.0]$ &Emergency Brake\\
	\bottomrule
	\end{tabular}	
\end{table}

The reward function consists of three components: the speed maintenance reward, collision penalty, and driving-too-slow penalty. The speed maintenance function in \eqref{eq:SHF} encourages the agent to drive the vehicle at 30 m/s. Note that the corresponding sensors attached to the ego vehicle return the speed and collision measurements for reward computation. These measurements are not revealed to the agent when devising controls. 
\begin{align}
	F_{SMR}(t)=\left\{
	\begin{array}{rcl}
	-0.005, & & {V(t)<0}\\
	\frac{1}{9}V(t)^2, & & {0\leq V(t) \leq 30}\\
	5(50-V(t)), & & {30< V(t) \leq 50}\\
	-2(V(t)-50)^2, & & {50< V(t)}
	\end{array} \right.,\label{eq:SHF}
\end{align}

To penalize the agent for colliding with other objects, we introduce the collision penalty function $F_{CP}$ defined in \eqref{eq:E_control}. $F_{CP}$ is a terminate reward function. The terminal reward depends on 1) whether the agent completes the task (the episode will be terminated and reset when collisions happen); 2)  how many steps the ego vehicle runs out of the lane in an episode. Denote by $t_{\text{terminate}}\leq H $ the terminal time step and by $N_{\text{out}}$ the number of time steps when the ego vehicle is out of the lane. Let $\delta(\cdot)$ be the Dirac function, and $F_{CP}$ is given by  
\begin{align}
	F_{CP}=&
	-100+({t}_{\text{terminate}}-H)\nonumber\\
	&+100\delta(t_{\text{terminate}}-H)-0.1N_{\text{out}},
	\label{eq:E_control}
\end{align}
penalizing the agent for not completing the task and running out of the lane. 

The agent is said to drive too slowly if the speed is lower than 0.5 m/s in two consecutive steps. Similar to the collision penalty, we introduce a driving-too-slowly(DTS) penalty as a terminal cost. Denote by $N_{slow}$ the number of steps when the vehicle runs too slowly, and the DTS penalty is given by
\begin{align}
	F_{DTS}=-0.005\times\max(50,10^{-7}\times N_{slow}).
	\label{eq:CS_control}
\end{align}

The horizon length is 1200 time steps, and the interval for adjacent frames is 0.05 seconds. The discount factor of all experiments is 0.99.

\paragraph{Dynamic Weather Setup}To realize dynamic weather conditions, we create a function $W(t)$ with respect to time defined in the following. To randomize the experiments, the initial value $W(0)$ is uniformly sampled from $[-150, 100]$ (subject to different random seeds in repeated experiemnts). To align time-varying weather with the built-in timeclock in CARLA,  we first introduce a weather update frequency $1/\Delta T, \Delta T>0$.  The whole epsidoe is evely divided into $n$ intervals: $[0,\Delta T), [\Delta T, 2\Delta T), \ldots, [(n-1)\Delta T, n\Delta T)$. Then, for $t\in [k\Delta T, (k+1)\Delta T), k>1$, the function $W(t)$ is given by 
\begin{align*}
    W(t)=\frac{25}{12}\cdot|\mbox{Mod}(1.3k\Delta T+W(0), 250)-125)|-150,
\end{align*}
and $W(t)=W(0)$, for $t\in[0,\Delta T)$.

Note that the hidden environment mode $z(t)$ consists of three weather parameters: clouds, rain, and puddles. Denote by $\operatorname{Clouds}(t), \operatorname{Rain}(t)$, and $\operatorname{Puddles}(t)$ the three parameters at time $t$, i.e., $z(t)=[\operatorname{Clouds}(t), \operatorname{Rain}(t), \operatorname{Puddles}(t)]$. Based on the function $W(t)$, these parameters are given by 
$$\left\{
\begin{aligned}
&\operatorname{Clouds}(t):=\mbox{Clip}(W(t)+40, 0,90), \\
&\operatorname{Rain}(t):=\mbox{Clip}(W(t), 0,80),\\
&\operatorname{Puddles}(t):=\mbox{Clip}(W(t)+d(t), 0,75),
\end{aligned}\right.
$$
where $d(t)$ is defined as the following: for $t\in [k\Delta T, (k+1)\Delta T), k\geq 0$,  
\begin{align*}
d(t) =
\begin{cases} 
-10,  & \mbox{if } W((k+1)\Delta T)\geq W(k\Delta T), \\
90, & \mbox{otherwise}.
\end{cases}
\end{align*}
$d(t)$ is a translation applied to $W(t)$ in the definition of $\operatorname{Puddles}(t)$ to preserve the causal relationship between rain and puddles in the simulation: puddles are caused by rain and appear after the rain starts.





\section{Training Details}\label{app:training}
 \paragraph{The Actro-Critic Policy Model} Our policy models under three weather conditions: cloudy, rainy, and dynamic changing weather, are trained via the Asynchronous Advantage Actor-Critic algorithm (A3C) with Adam optimizer mentioned in~\cite{Palanisamy:2018:HIA:3285236}. The learning rate begins at $1\times 10^{-4}$, and the policy gradient update is performed every 10 steps. The reward in each step is clipped to -1 or +1, and the  entropy regularized method \cite{entropy_bonus} is used. Once episode rewards stabilize, the learning rate will be changed to $1\times 10^{-5}$ and $1\times 10^{-6}$. 

\begin{figure}
    \centering
    \includegraphics[width=0.5\textwidth]{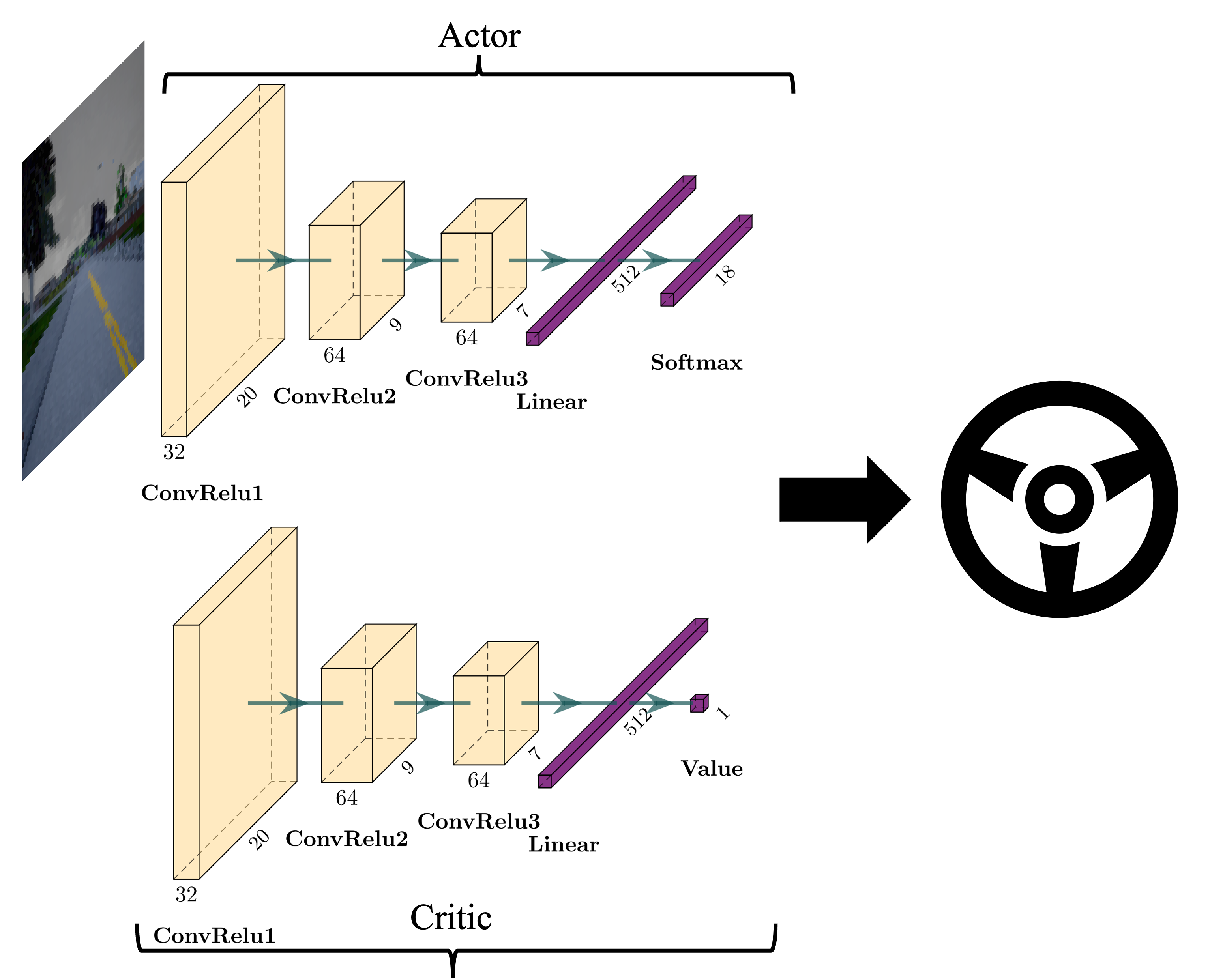}
    \caption{Network structures of the actor and the critic. The perception and decision-making module are incorporated into a single policy model that outputs control commands when fed with raw images.}
    \label{fig:ac}
\end{figure}

The policy network consists of two parts: the Actor and the Critic. They incorporate three convolution layers that use Rectified Linear Units (ReLu) as activation functions. The input is a $1\times 6 \times 84 \times 84$ dimension image data transformed from two $80\times80\times3$ RGB images. A linear layer is appended to convolution layers. The softmax layer serves as the last layer in the Actor and the Critic. The schematic diagram of the model structure is shown in \Cref{fig:ac}.

\paragraph{Image Classifier} Our image classifier is based on a residual neural network~\cite{resnet} with two residual blocks and one linear output layer. The input is the current camera image, and the output is the probability of that image being of the rainy type. The data sets (training, validation, and testing) include RGB camera images under the cloudy and rainy conditions in the CARLA world. 

Every image input is paired with a three-dimensional vector $z$ (the mode), with entries being the values of three weather parameters: cloudiness, rain, and puddles, respectively. Denote the corresponding mode of an image by $z_{img}$. Let $z_{r}$, and $z_{c}$ be the parameter vectors provided by the simulator in the default rainy and cloudy conditions, respectively. Then, the true label $i$ of an image is given by 
\begin{align*}
	\text{Label}=\left\{\begin{array}{ll}
		1,& \text{if }  \|z_{img}-z_{r}\|<\|z_{img}-z_{c}\|, \\
		0, & \text{otherwise}.
	\end{array}\right.
\end{align*} 

 We use One-Cycle-Learning-Rate~\cite{OneCycleRL} in the classifier training, with the max learning rate $1\times 10^{-3}$ in the first 120K iterations of training and $1\times 10^{-6}$ in the later 600K iterations for fine-tune training. The test accuracy of the classifier is $91.7\%$.
 
\section{The Tradeoff of Lookahead Horizon}
\label{app:tradeoff}
The impact of the lookahead horizon $K$ on the online adaptation is due to a tradeoff between variance reduction and belief calibration. If $K$ is small, the sample-based estimate of $\E[\sum_{k=0}^{K-1} r(s_{t+k},a_{t+k})]$ would incur a large variance, which is then propagated to the gradient computation $\hat{g}$ in \eqref{eq:opa}. Hence, the resulting approximate also exhibits a large variance. On the other hand, when $K$ is large, the agent may look into the future under the wrong belief since the agent believes that the environment is stationary for the future $K$ steps. In this case, the obtained policy may not be able to adapt promptly to the changing environment. \Cref{tab:Step_table} summarizes the average rewards under three lookahead horizons $K=5,10,20$. As we can see from the table, the average rewards under $K=5$ exhibit the largest variances across all weather conditions in comparison with their counterparts under $K=10,20$. Even though $K=20$ achieves performance in stationary environments (i.e., cloudy and rainy) comparable to that under $K=10$, we choose $K=10$ in our experiments as it balances variance reduction and quick belief calibration, leading to the highest rewards in the dynamic weather. 
\begin{table}
\small
    \centering
    \begin{tabular}{cccc}
    \toprule
    K & Cloudy & Rainy & Dynamic\\
     \midrule
    {5} &$113.16\pm 47.80$ &$-156.36\pm 60.83$ &$109.72\pm 57.45$\\
    \midrule
     {10}   &$159.09\pm 38.41$ &$-145.75\pm 47.64$ &$159.89\pm 51.35$\\
     \midrule
    {20} &$150.41\pm 43.68$ &$-136.72\pm 55.33$ &$124.03\pm 52.36$\\
    \bottomrule
    \end{tabular}
    \caption{The average cumulative rewards and associated standard deviations under different lookahead horizons $K$ in three weather conditions.}
    \label{tab:Step_table}
\end{table}
\end{appendices}
\end{document}